\documentclass[10pt,twocolumn,letterpaper]{article}

\usepackage[pagenumbers]{cvpr} 
\usepackage{multirow}
\usepackage{adjustbox}
\usepackage[dvipsnames]{xcolor}

\usepackage{pifont}
\usepackage{array}
\usepackage{booktabs}
\usepackage{tcolorbox}
\usepackage{algorithm2e}
\usepackage{balance}
\usepackage{amsmath}
\newtheorem{theorem}{Theorem}

\usepackage{ulem}
\usepackage{subcaption}
\usepackage{graphicx}
\newtheorem{corollary}{Corollary}

\usepackage{xcolor}
\usepackage{colortbl}
\definecolor{cvprblue}{rgb}{0.21,0.49,0.74}
\usepackage[pagebackref,breaklinks,colorlinks,citecolor=cvprblue]{hyperref}

\def\paperID{8905} 
\def\confName{CVPR}
\def\confYear{2025}

\title{Large Language Models for Lossless Image Compression:\\ Next-Pixel Prediction in Language Space is All You Need}

\author{
Kecheng Chen\textsuperscript{1}, 
Pingping Zhang\textsuperscript{2}, 
Hui Liu\textsuperscript{1}, 
Jie Liu\textsuperscript{1}, 
Yibing Liu\textsuperscript{2}, \\
Jiaxin Huang\textsuperscript{3}, 
Shiqi Wang\textsuperscript{2}, 
Hong Yan\textsuperscript{1}, 
Haoliang Li\textsuperscript{1} \\
\textsuperscript{1}Department of Electrical Engineering and \\the Centre for Intelligent Multidimensional Data Analysis (CIMDA), City University of Hong Kong \\
\textsuperscript{2}Department of Computer Science, City University of Hong Kong \\
\textsuperscript{3}Mohamed bin Zayed University of Artificial Intelligence \\
}

\begin{document}
\maketitle
\begin{abstract}
We have recently witnessed that ``Intelligence" and `` Compression" are the two sides of the same coin, where the language large model (LLM) with unprecedented intelligence is a general-purpose lossless compressor for various data modalities. This attribute particularly appeals to the lossless image compression community, given the increasing need to compress high-resolution images in the current streaming media era. Consequently, a spontaneous envision emerges: Can the compression performance of the LLM elevate lossless image compression to new heights? However, our findings indicate that the naive application of LLM-based lossless image compressors suffers from a considerable performance gap compared with existing state-of-the-art (SOTA) codecs on common benchmark datasets. In light of this,  we are dedicated to fulfilling the unprecedented intelligence (compression) capacity of the LLM for lossless image compression tasks, thereby bridging the gap between theoretical and practical compression performance. Specifically, we propose P$^{2}$-LLM, a next-pixel prediction-based LLM, which integrates various elaborated insights and methodologies, \textit{e.g.,} pixel-level priors, the in-context ability of LLM, and a pixel-level semantic preservation strategy, to enhance the understanding capacity of pixel sequences for better next-pixel predictions. Extensive experiments on benchmark datasets demonstrate that P$^{2}$-LLM can beat SOTA classical and learned codecs. 
\end{abstract}
\section{Introduction}
\label{sec:intro}
Recently, \citet{deletang2024language} have uncovered that a large language model (LLM), pre-trained on massive text corpora, can achieve competitive lossless compression rates across text, audio, and image modalities. This perspective derives from the so-called philosophy, \textit{``Intelligence" and ``Compression" are two sides of the same coin}~\citep{mackay2003information}. Theoretically, minimizing log-loss for next-token prediction in the LLM is equivalent to optimizing a lossless compression objective, positioning the LLM as a \textit{general-purpose} compressor for any modality~\citep{heurtel2024compression}. This insight is particularly compelling for the lossless image compression community, where the need for more effective compression methods has become increasingly critical in the era of streaming media~\citep{rahman2019lossless}. As advanced LLMs' intelligence gradually outperforms humans in various applications~\citep{hu2024can}, a spontaneous envision emerges, \textit{i.e.,} \textit{Can the compression performance of the LLM elevate lossless image compression to new heights?} If the answer is affirmative, the lossless image compression community will benefit steadily from the progress in LLM techniques, since \citet{huang2024compression} recently reveal that a linear growth relationship between LLM's compression performance and intelligence holds. 

\begin{figure}[!t]
    \centering
    \includegraphics[width=0.85\linewidth]{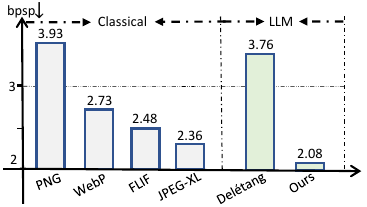}
    \caption{Comparison of different lossless image compressors for bit-per-subpixel (bpsp$\downarrow$) on CLIC.m dataset. Classical compressors include PNG, WebP, FLIF, and JPEG-XL.}
\label{performance}
\vspace{-0.2cm}
\end{figure}

However, achieving this roadmap is not straightforward. The current LLM-based compressor~\citep{deletang2024language} primarily showcases the methodology and corresponding compression results on grayscale images, leaving it unclear how to extend the lossless image compression capabilities of the LLM to more widely-used images, such as RGB images.
As depicted in Figure \ref{performance}, the direct application of the existing LLM-based compressor to benchmark datasets comprising RGB images reveals a significant performance disparity compared with state-of-the-art (SOTA) classical lossless codecs, for instance, JPEG-XL~\citep{alakuijala2019jpeg}.


To unlock the potentially unprecedented intelligence (compression) capacity of the LLM and bridge the gap between theoretical and practical compression performance,  we carefully analyze three potential limitations of existing LLM-based lossless image compressor~\citep{deletang2024language} when applied to widely-used images. First, compared with next-token (pixel) prediction for grayscale images, modeling the highly nonlinear and long-range correlations of RGB images, where each pixel comprises three subpixels, is more sophisticated. The existing method~\citep{deletang2024language} lacks an effective mechanism to address this challenge. Second, \citet{deletang2024language} impose proxy tokens (\textit{i.e.,} ASCII characters) to represent each pixel value in language space, which would discard the original pixel-level semantic context, impairing the LLM’s ability to understand images through their numerical pixel values in language space. Third, the naive application of the pre-trained LLM to next-pixel prediction tasks may inherently result in suboptimal performance, as general foundation models typically lack the domain-specific knowledge required for customized tasks~\citep{hu2021lora}.

To address these challenges, we aim to reformulate the overall framework of LLM-based lossless image compression into a new one, which can unlock the inherent intelligence (compression) ability of the LLM to achieve comparative or better compression performance compared with SOTA lossless image codecs. With this goal in mind, our primary motivation is to boost LLM's capacity to comprehend highly complex and long-range correlated pixel sequences in language space, thereby improving next-pixel prediction accuracy, which directly correlates with a better compression ratio~\citep{zhu2024language}. Specifically, we first propose to leverage pixel-level priors (\textit{e.g.,} intra-pixel inter-channel correlation and local self-similarity) and the in-context ability of the LLM to facilitate the understanding of complex RGB pixel sequences. To this end, we integrate these functionalities into a pixel prediction chat template. Second, instead of using proxy tokens, we propose a two-step lossless pixel tokenization strategy that maximizes pixel-level semantic preservation for LLM context understanding, where each subpixel is treated as a ``word" that corresponds to a numerical representation in the token dictionary. This is motivated by recent findings, \textit{i.e.,} \citet{zhu2024languagemodelsknowvalue} uncover that \textit{LLMs can understand the value of numbers such as the numerical relationship (\textit{e.g.,} one number is smaller or larger than another)}. Finally, we employ a low-rank adaptation (LoRA)-based fine-tuning strategy~\citep{hu2021lora}, which efficiently and effectively enhances the LLM's understanding capacity of LLM for this customized pixel prediction task. The overall framework is termed as P$^{2}$-LLM, \textit{i.e.,} next-\underline{p}ixel \underline{p}rediction-based \underline{LLM}. Our contributions can be summarized as three-fold:

\begin{enumerate}
\item[$\bullet$] We aim to fully unlock LLM's unprecedented intelligence (compression) capacity for the lossless image compression task. This perspective bridges the gap between theoretical and practical compression performance for LLM, potentially opening new avenues as LLM intelligence continues to evolve in the future.

\item[$\bullet$] We propose P$^{2}$-LLM, a next-pixel prediction-based LLM that integrates various elaborated methodologies, \textit{e.g.,} pixel-level priors, the in-context ability of LLM, and pixel-level semantic preservation strategy. These elements collectively enhance the LLM’s capacity to comprehend pixel sequences for next-pixel predictions.

\item[$\bullet$] The proposed P$^{2}$-LLM significantly improves the lossless image compression rate of existing LLM-based compressors. Although P$^{2}$-LLM has no visual-perception architecture, it can beat SOTA classical and learned codecs, which showcase \textit{the next-pixel prediction of LLM in language space may be all you need}. Notably, on CLIC.m and Kodak datasets, P$^{2}$-LLM achieves 2.08 and 2.83 bit-per-subpixel (bpsp$\downarrow$), respectively, suppressing the current best approach (DLPR~\citep{bai2024deep}), which records 2.16 and 2.86 bpsp, respectively.

\end{enumerate}

\begin{figure*}[!h]
    \centering
    \includegraphics[width=\linewidth]{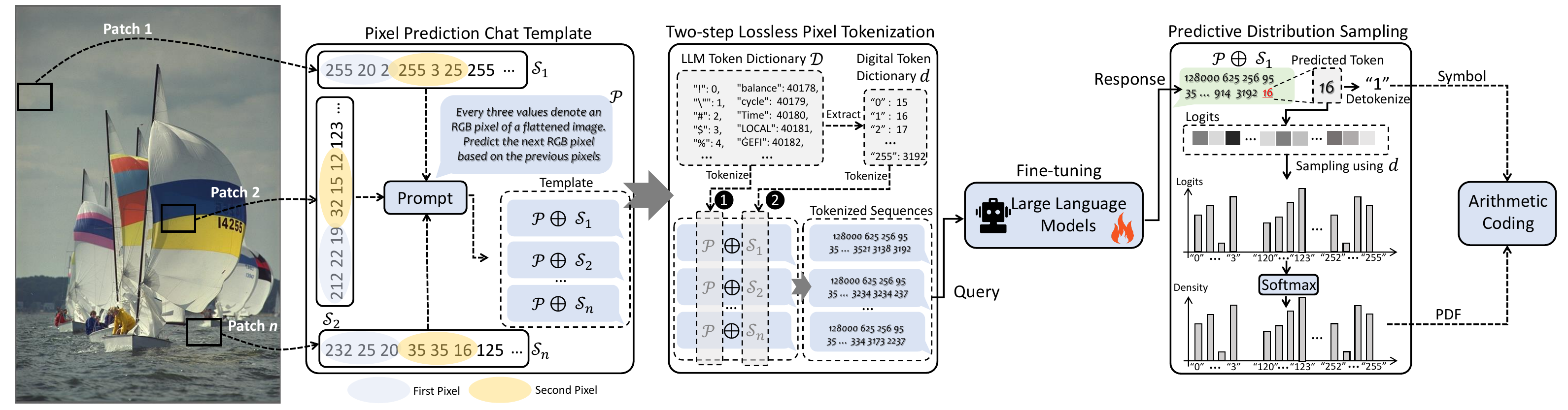}
    \caption{The framework of the proposed P$^{2}$-LLM, including Pixel Prediction Chat Template for the pixel-level priors and in-context integration in sec. \ref{3.1}, Two-step Lossless Pixel Tokenization for pixel-level semantic preservation in sec. \ref{3.2}, Predictive Distribution Sampling for scalable probability representation of encoded symbols in sec. \ref{3.3}, and Fine-tuning to boost the understanding capacity of pixel sequences in sec. \ref{3.4}. You may zoom in for a better view. }
    \label{framework}
\end{figure*}
\section{Related works}
\subsection{Lossless Image Compression} 
Traditional lossless image codecs, such as PNG~\citep{boutell1997png} and JPEG-XL~\citep{alakuijala2019jpeg}, operate by employing manual pipelines to diminish the redundancy of images for compression. However, due to the optimized difficulty of traditional codes, the performance gradually bounds with little increase. Thus, the learned image compression (LIC) approaches aim to mitigate such issues by an end-to-end learning framework~\citep{bai2024deep}. Usually, LIC methods encompass two steps, including 1) statistical modeling of a given image using deep generative models and 2) encoding the given image into the bitstreams using arithmetic coding. Herein, modeling accurate and generalizable statistics of the given image is the key component, where various generative models are used as follows. 1) Autoregressive models, such as PixelRNN and PixelCNN~\citep{van2016pixel}, which forecast pixel distributions based on conditional dependencies with previously acquired pixels via masked convolutions. 2) Flow models, \textit{e.g.,} iVPF~\citep{zhang2021ivpf} and iFlow~\citep{zhang2021iflow}, leverage invertible transforms to simplify latent distributions for efficient entropy coding. 3) Variational Auto-Encoder models, such as L3C~\citep{mentzer2019practical}, which utilize variational architectures to model image distributions.

In this paper, LLM-based compressors belong to autoregressive models, but there are no visual-perception components (\textit{e.g.,} masked convolutions). More importantly, instead of perceiving images directly, LLM-based compressors model the statistics of the image in the language space by discretizing each pixel to language tokens. 
\subsection{Large Language Models for Compression}
The large language model (LLM) has performed surprisingly well in natural language processing~\citep{wu2023next} and computer vision tasks~\citep{yao2024survey}, due to its accurate next-token prediction capacity. For example, many challenging applications, \textit{e.g.,} machine translation~\citep{feng2024improving}, language understanding~\citep{jiang2024look}, and affective forecasting~\citep{zhang2024affective} are intensively solved by LLM. 

Recently, \citet{deletang2024language} demonstrated that language modeling is compression, as log-loss minimization for the next-token prediction of LLM is equivalent to optimizing a lossless compression objective, which enables the LLM as a \textit{general-purpose} compressor for any modality~\citep{heurtel2024compression}. They showcased that LLM-based compressors can beat some classical codecs (\textit{e.g.,} PNG) for grayscale images. Moreover, recent literature also implies the linear growth relation between compression performance and LLM's intelligence~\citep{huang2024compression}. These insights motivate the lossless image compression community to investigate the unprecedented intelligence of LLM in more common images, \textit{e.g.,} RGB images. Although it is straightforward to extend \citet{deletang2024language}' approach to RGB images in a channel-independent manner, \textit{i.e.,} each channel of RGB images is regarded as a grayscale image, such strategy suffers from many limitations as discussed before. Thus, we are dedicated to fulfilling LLM's unprecedented intelligence (compression) capacity for the lossless image compression task.

\section{Methodology}
\textbf{Overall.} From a lossless compression perspective, the proposed P$^{2}$-LLM (as depicted in Figure \ref{framework}) aims to render an accurate probability representation of each encoded symbol (\textit{i.e.,} the (sub)pixel) for arithmetic coding. Note that arithmetic coding is acknowledged to be optimal for coding length, where the
overall compression performance depends on the abilities of the probabilistic model~\citep{deletang2024language}. To this end,  we focus on unlocking the unprecedented reasoning capacity of LLM to understand pixel sequence for better next-pixel predictions with accurate probability representations.


\subsection{Pixel-level Prior and In-context Integration }\label{3.1}
\begin{figure}[!t]
    \centering
    \includegraphics[width=\linewidth]{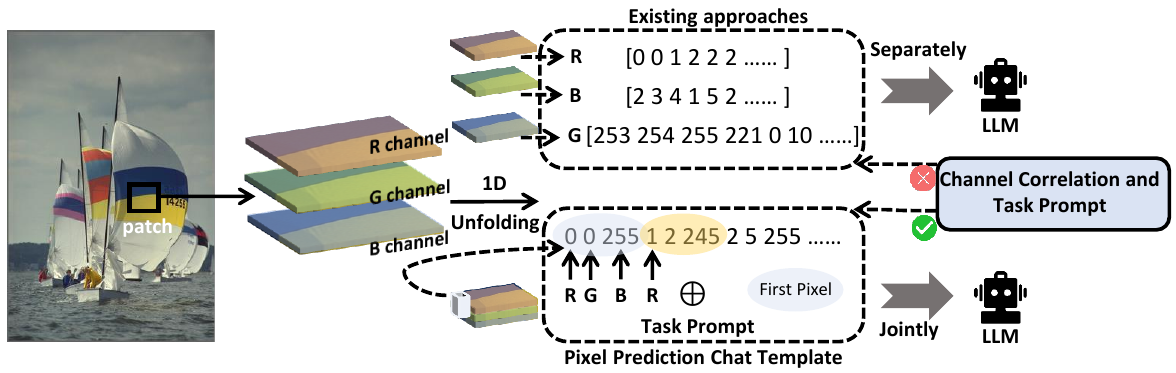}
    \caption{Comparison of different input sequences for pixel prediction. The existing approach is from \citet{deletang2024language}. We process each patch of RGB images in a channel-joint manner. }
    \label{fig:flatten}
    \vspace{-0.2cm}
\end{figure}

In this paper, 
we assume that LLMs can capture implicit structured information and patterns to conduct next-pixel prediction. Such capacity derives from the unprecedented intelligence of the LLM learned from the massive text corpus~\citep{deletang2024language}. However, existing results on benchmark datasets demonstrate that the pre-trained LLM are still behind SOTA codecs with a significant gap.  We argue that this performance gap may derive from two aspects:
\begin{itemize}
\item \textbf{Without Pixel-level Priors:} Previous literature usually realizes accurate next-pixel prediction based on fruitful pixel-level priors, \textit{e.g.,} intra-pixel inter-channel correlation ~\citep{van2016pixel,salimans2017pixelcnn} and local self-similarity~\citep{zhang2023considering,wewer2023simnp} between pixels. However, existing LLM-based compressors fail to leverage these pixel-level priors to reason the next prediction, as they either neglect the inter-channel correlation~\citep{li2024understanding} by channel-independent processing or discard channel information by graying~\citep{deletang2024language}.  
\item \textbf{Without Leveraging In-context Learning of LLMs:} Quite a few works~\citep{dong2024survey} have demonstrated that in-context learning with well-supported prompts can help the LLM understand the task, leading to more accurate predictions. However, existing LLM-based compressors simply input the pixel sequence without motivating potential next-pixel prediction ability under a specific context. 
\end{itemize}
To address the aforementioned limitations, we propose to integrate pixel-level priors and the in-context ability of LLM to enhance the ability to reason the next pixel. To this end, we design a customized pixel prediction chat template to integrate these functions into one. Specifically, given an RGB image $\mathbf{X} \in \mathbb{R}^{W\times H \times 3}$ where $W$ and $H$ denotes the spatial resolution, $\mathbf{X}$ is first flattened into a 1D sequence with $W\times H$ pixels, \textit{i.e.,} $\mathbf{X}^{'} =$
\begin{small}
    \begin{equation}
     \mathrm{flatten}(\mathbf{X}) = [ \mathbf{x}_{1}, \mathbf{x}_{2},\cdots, \mathbf{x}_{W\times H}], \mathbf{x}_{i} = \{x_{i}^{R},x_{i}^{G},x_{i}^{B}\},
\end{equation}
\end{small}where $\mathbf{x}_{i}$ denotes a pixel that consists of three subpixels $x_{i}^{R},x_{i}^{G},x_{i}^{B}$ for red, green, and blue channels. As illustrated in Figure \ref{fig:flatten}, sequential pixels with inter-channel correlation can potentially boost the understanding ability of LLM for spatial relationships, \textit{e.g.,} local self-similarity between continues pixels and color consistency in a specific range. Instead, the relationship between subpixels in a channel-independent manner can be easily disturbed by various interferences, \textit{e.g.,} the noise and the downsampling. 

Meanwhile, we introduce a task prompt $\mathcal{P}$ to motivate the in-context understanding ability of LLM. An example task prompt $\mathcal{P}$ can be presented as follows:
\begin{tcolorbox}
\uwave{Every three values denote an 
RGB pixel}\footnote{Instruct the relationship between sequential values.} of \uwave{a flattened image}\footnote{Clarify the data format.}. 
\uwave{Predict the next RGB pixel 
based on the previous pixels.}\footnote{Clarify the reasoning task.}
\end{tcolorbox}
We can observe from the task prompt that effective instructions, \textit{e.g.,} the relationship between sequential values and the reasoning task, are provided to potentially enhance the reasoning capacity of LLM for next-pixel predictions.
Formally, our proposed pixel prediction chat template can be represented as follows:
\begin{equation}
    \mathbf{S} =  \mathcal{P} \oplus \mathbf{X}^{'},
\end{equation}
where $\oplus$ denotes the concatenation operation. 
Thus,  the conditional probability of a
symbol $x_{3i+k}$ ($k$ is 1, 2, and 3 for R, G, and B channel, respectively) given previous $i$ pixels $\mathbf{x}_{1:i} = x_{1:3i}$ and the task prompt $\mathcal{P}$ can be defined as 
\begin{equation}
    \rho(x_{3i+k}|x_{1:3i}, \mathcal{P}) = \rho(x_{1:3i+k}|\mathcal{P})/\rho(x_{1:3i}|\mathcal{P}).
\end{equation}

\textbf{Theoretical Analysis.} Different from channel-independent prediction, we use the LLM to conduct a sequential prediction of three channels of a pixel by leveraging inter-channel correlation. To demonstrate the rationality of such a strategy, we first provide the universally-defined prediction theory by neural network-based amortization of Solomonoff induction~\citep{salimans2017pixelcnn}, as follows, 
\begin{theorem}\citep{li2024understanding,grau2024learning} \label{theory1}
    For any parametric meta-learning model $f_{\theta}$ like the decoder-only large models, if $f_{\theta}$ is fully trained by log-loss function and consider an infinite sequence $\omega$ of events over a finite alphabet,
the optimum posterior distribution $\mu$ of $\omega_{i+1}$ given $\omega_{1:i}$ can be obtained, \textit{i.e.,}
\begin{equation}
    \lim_{i\rightarrow\infty} (f_{\theta}(\omega_{i+1}|\omega_{1:i})-\mu(\omega_{i+1}|\omega_{1:i}))=0 
\end{equation}
\end{theorem}
We extend the result in Theorem \ref{theory1} to our setting, \textit{i.e.,} a sequential prediction of three subpixels of a pixel, to clarify a similar optimum posterior distribution can be derived:  
\begin{corollary}
The optimum posterior distribution $\mu$ of $x_{i+1}^{R}$,$x_{i+1}^{G}$, and $x_{i+1}^{B}$ from a perspective of joint distribution, given previous $i$ pixels $\mathbf{x}_{1:i}$,  can be obtained as follows
\begin{small}
    \begin{equation}
        \lim_{i\rightarrow\infty} (f_{\theta}(x_{i+1}^{R},x_{i+1}^{G},x_{i+1}^{B}|\mathbf{x}_{1:i})-\mu(x_{i+1}^{R},x_{i+1}^{G},x_{i+1}^{B}|\mathbf{x}_{1:i}))=0 \nonumber,
    \end{equation}
\end{small}
\label{coro}where optimum posterior distribution results in the smallest coding length for arithmetic coding.
\end{corollary}
\textbf{\textit{Proof.}}
First, we can decompose the joint distribution $\mu(x_{i+1}^{R},x_{i+1}^{G},x_{i+1}^{B}|\mathbf{x}_{1:i}))$ using chain rule:
\begin{small}
    \begin{equation}
      \mu(x_{i+1}^{R} | \mathbf{x}_{1:i}) \cdot \mu(x_{i+1}^{G} |\mathbf{x}_{1:i}, x_{i+1}^{R}) \cdot \mu(x_{i+1}^{R} | \mathbf{x}_{1:i}, x_{i+1}^{R}, x_{i+1}^{G}). \label{mu} \nonumber
\end{equation}
\end{small}Similarly, for $f_{\theta} =$ 
\begin{small}
\begin{equation}
    f_{\theta}(x_{i+1}^{R} | \mathbf{x}_{1:i}) \cdot f_{\theta}(x_{i+1}^{G} |\mathbf{x}_{1:i}, x_{i+1}^{R}) \cdot f_{\theta}(x_{i+1}^{B} | \mathbf{x}_{1:i}, x_{i+1}^{R}, x_{i+1}^{G}). \nonumber \label{f}
\end{equation}
\end{small}By recalling Theorem \ref{theory1}, if each pair of conditional distribution converges independently, a fully trained $f_{\theta}$ is a must. This means $f_{\theta}$  has to capture correlations using previous subpixels of the current pixel and previous pixels as the condition, ensuring accurate predictions. However, such domain-specific capacity cannot be guaranteed strictly by pre-trained $f_{\theta}$. Thus, we assume that such a fully trained decoder-only model can be obtained by fine-tuning pre-trained $f_{\theta}$ to $\hat{f}_{\theta}$. Then, 
\begin{small}
    \begin{equation}
    \lim_{i\rightarrow\infty} (\hat{f}_{\theta}(x_{i+1}^{R}|\mathbf{x}_{1:i})-\mu(x_{i+1}^{R}|\mathbf{x}_{1:i})=0 \label{R converge} \nonumber
\end{equation}
\end{small}
\begin{small}
    \begin{equation}
    \lim_{i\rightarrow\infty} (\hat{f}_{\theta}(x_{i+1}^{G}|\mathbf{x}_{1:i},x_{i+1}^{R})-\mu(x_{i+1}^{G}|\mathbf{x}_{1:i},x_{i+1}^{R}))=0 \label{G converge} \nonumber
\end{equation}
\end{small}
\begin{small}
    \begin{equation}
    \lim_{i\rightarrow\infty} (\hat{f}_{\theta}(x_{i+1}^{B}|\mathbf{x}_{1:i},x_{i+1}^{R},x_{i+1}^{G})-\mu(x_{i+1}^{B}|\mathbf{x}_{1:i},x_{i+1}^{R},x_{i+1}^{G}))=0 \label{B converge} \nonumber
\end{equation}
\end{small}As the convergence for each component implies the convergence of the product of these components due to the properties of limits and continuity, Cor. \ref{coro} holds. However,
it is necessary to fine-tune $f_{\theta}$ to well model conditional distributions and related correlations, otherwise suboptimal posterior distributions will be, due to suboptimal convergence.$\square$

Overall, Cor. \ref{coro} implies that our channel-joint training can encourage the LLM to implicitly learn a joint distribution over subpixels of a pixel and capture correlations of conditional distributions for optimum posterior distribution. Compared with channel-independent-based posterior distributions, such modeling will result in more robust and accurate representations as discussed by \citet{salimans2017pixelcnn} (which rely on explicitly parameterized modeling). Meanwhile, fine-tuning (as described in sec. \ref{3.4}) is indispensable to realize optimum posterior distribution. 


\begin{figure}[!t]
    \centering
    \begin{subfigure}[b]{0.5\textwidth}
        \centering
        \includegraphics[width=\textwidth]{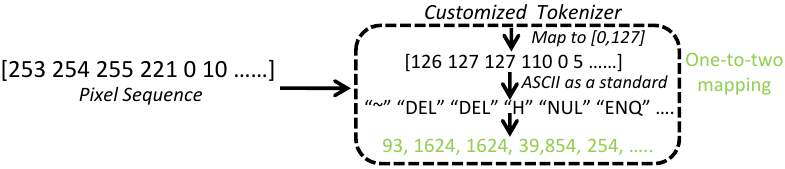}
        \caption{}
        \label{fig:sub1}
    \end{subfigure}
    \hfill
    \begin{subfigure}[b]{0.5\textwidth}
        \centering
        \includegraphics[width=\textwidth]{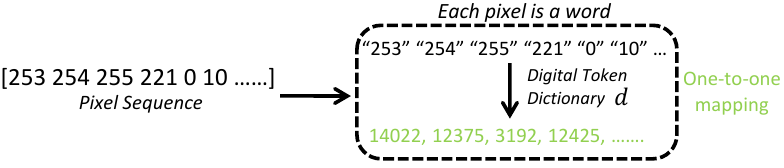}
        \caption{}
        \label{fig:sub2}
    \end{subfigure}
    \caption{(a) Customized Tokenizer~\citep{deletang2024language} (b) Lossless Image Tokenizer used by ours. Note that the subpixel values, 254 and 255, correspond to the same token ID (\textit{i.e.,} 1624) in (a). }
    \label{fig:tokenization}
\end{figure}
\begin{figure*}[!h]
    \centering
    \includegraphics[width=0.9\linewidth]{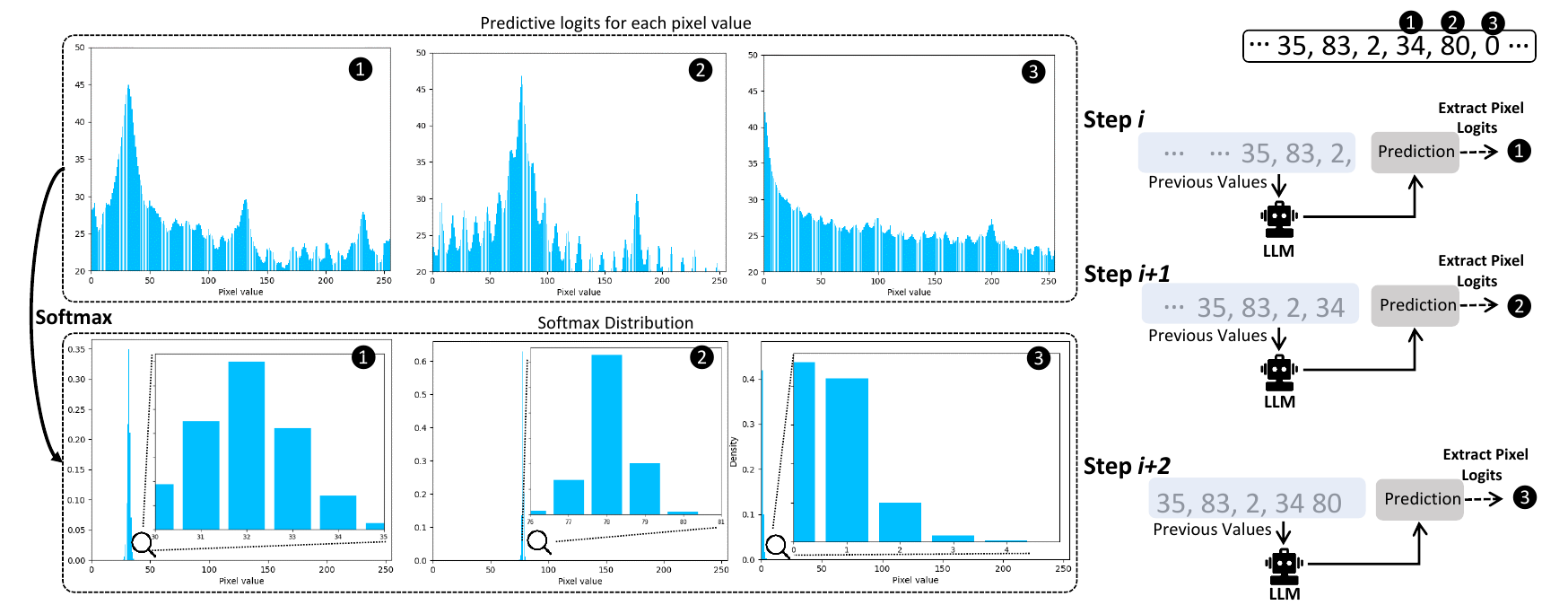}
    \caption{Visualization of predictive distribution sampling for reasoning three subpixels using LLM. You may zoom in for a better view.}
    \label{fig:sampling}
\end{figure*}
\subsection{Two-step Lossless Pixel Tokenization}\label{3.2}
The tokenizer is an important component in bridging the original semantic space and discrete language representation used by LLMs. Recent progresses~\citep{ali2023tokenizer} highlight the tokenizer choice and corresponding token representations can significantly impact the LLM's downstream performance and reasoning ability. Motivated by this, existing LLM-based compressors may be suboptimal, as follows:
\begin{itemize}
    \item \textbf{Without One-to-One Mapping:} 
To ensure lossless compression, the tokenizer must enable a one-to-one mapping between the pixel (subpixel for RGB images) value and the token representation. However, as shown in Figure \ref{fig:tokenization}, \citet{deletang2024language} conduct a data mapping from the original pixel representation to the range [0,127], as ASCII charterers are encoded in this range. Thus, such preprocessing results in \textit{one-to-two} mapping with information loss. Although they append lost bits to the end of the compressed sequence, the compression ratio decreases.
\item \textbf{Without Pixel-level Semantic Context:} Intuitively, if the discrete token representation maintains the original pixel-level semantic context,  those pixel-level priors will still be utilized by LLMs. However, as illustrated in Figure \ref{fig:tokenization}, existing approaches~\citep{deletang2024language} use the customized tokenizer with ASCII characters but fail to do so. For example, the pixel values \texttt{126}, \texttt{127}, and \texttt{0} are represented as ``\texttt{\~}", ``\texttt{DEL}", and ``\texttt{NUL}", respectively. The closer relationship between \texttt{126} and \texttt{127} violates.
\end{itemize}

To tackle the aforementioned limitations, we propose a two-step lossless pixel tokenization strategy. In a nutshell, our solution derives from the on-hand numerical understanding ability of LLM (\textit{e.g.,} one number is smaller or larger than another)  as discussed by \citet{zhu2024languagemodelsknowvalue}. Motivated by this finding,  we propose to treat each subpixel as a word that corresponds to a numerical representation in the token dictionary, \textit{i.e.,} \texttt{127} $\rightarrow$ ``\texttt{127}".
By doing so, we can 
achieve a one-to-one mapping from the pixel value to the token representation, as each digital word from ``\texttt{0}" to ``\texttt{255}" has a unique token ID. More importantly, due to the preservation of pixel-level semantic context, LLM can understand the sequential pixel values for better next-pixel prediction. 

Specifically, the two-step lossless pixel tokenization framework includes 1) a widely used tokenization process and 2) a one-to-one matching process between digital words and digital tokens. Formally, given the tokenization function as $T(\cdot)$ and the corresponding token dictionary as $\mathcal{D}$, we first extract the digital words in the range of 0 to 255 and the corresponding token ID, as follows
    \begin{equation}
     d = \{ \texttt{str}(z): T(\texttt{str}(z),\mathcal{D}) \mid z \in \{0, 1, 2, \ldots, 255\} \},
\end{equation}where $d$ denotes the digital token dictionary and $\texttt{str}(\cdot)$ denotes a number-to-string conversion operator for each pixel value $z$. Furthermore, the task prompt $\mathcal{P}$ and input pixel sequence $\mathbf{X}^{'}$ are tokenized by two steps as follows:
     \begin{equation}
    \mathbf{S}^{'} = T(\mathcal{P},\mathcal{D}) \oplus\{ d(\texttt{str}(z_{j}))\}_{j=1}^{W\times H \times 3},
 \end{equation}where we look up the token representation of each pixel value using the obtained digital token dictionary $d$, which ensures the one-to-one mapping to avoid potential word splitting using tokenizer function $T(\cdot)$. 

\subsection{Predictive Distribution Sampling}\label{3.3}
 LLM can conduct next-token predictions to output the predictive softmax probability. However, only the pixel value prediction is required for lossless image compression. We therefore propose to sample the predictive logits before the softmax layer, based on the digital token dictionary $d$. Then, these sampled logits will be normalized to a probability distribution consisting of 256 probabilities proportional to the exponentials of the input numbers. Such probability distribution can be used for arithmetic coding. Compared with previous works, \textit{e.g.},~\citet{bai2024deep} that need to learn parameterized distributions (\textit{e.g.,} Gaussian mixture models), our sampled predictive distribution is parameter-free with maximal scalability and robustness in complex scenarios~\citep{broom2007parameter}.
 
 Formally, given the LLM as a next-token prediction function $\texttt{LLM}(\cdot)$ \textit{without} the softmax layer, the prediction is represented as $\mathbf{y} = \texttt{LLM}(\mathbf{S}^{'})$, where $\mathbf{y} \in \mathbb{R}^{1\times |\mathcal{D}|}$ consists of $|\mathcal{D}|$ predictive logits $\{y_{n}\}_{n=1}^{|\mathcal{D}|}$ for each token ID, \textit{i.e.,} $n$. By using the digital token dictionary $d$, the pixel-related predictive vector $\mathbf{y}_{c}$ can be represented as 
 \begin{small}
     \begin{equation}
     \mathbf{y}_{c} = \{ y_{c}^{z} = y_{n}  \mid n=d(\texttt{str}(z)), z \in \{0, 1, 2, \ldots, 255\} \},
 \end{equation}
 \end{small}where $\mathbf{y}_{c}\in \mathbb{R}^{1\times 256}$. As shown in Figure \ref{fig:sampling}, the predictive logtis between 0 and 255 are presented in the middle column. For each prediction, the peak of those logits is roughly around the encoded pixel value, which showcases that LLM can effectively predict the next pixel. Meanwhile, the visualized logits in Figure \ref{fig:sampling} demonstrate that LLMs can naturally assign more mass to most possible pixel values. Instead, previous methods~\citep{salimans2017pixelcnn} need to assign a higher probability to the edge values 0 and 255 in a handcrafted manner.
\begin{table}[!t]
    \centering
    \caption{The datasets for evaluation. The symbols $*$ and $\dag$ denote fine-tuning and testing datasets, respectively.}
    \renewcommand\arraystretch{0.85}
 \begin{adjustbox}{max width=\columnwidth}
	\begin{tabular}{cccc}
\hline
\textbf{Dataset}         & \textbf{Description} &  \textbf{\# Num.}  & \textbf{Avg. Resolution}\\ \hline 
\rowcolor{gray!15} DIV2K-Training$^{*}$   & Natural            & 800 &  1080$\times$2048   \\
Kodak$^{\dag}$   & Natural            & 24 &  576$\times$704   \\
\rowcolor{gray!15}DIV2K-Validation$^{\dag}$          & Natural &  100    &  1080$\times$2048  \\
SCID$^{\dag}$ & Screen Content              &  40   &     720$\times$1080 \\
\rowcolor{gray!15}CLIC.p$^{\dag}$ &   Natural           &   41   & 1080$\times$2048 \\
BRACS24$^{\dag}$           & Medical             & 24   &  1526$\times$2897\\
\rowcolor{gray!15} CLIC.m$^{\dag}$         & Natural             &  61  &   1080$\times$2048   \\ \hline   
\end{tabular}
\label{dataset}
\end{adjustbox}
\end{table} 
\begin{table*}[!h]
\renewcommand\arraystretch{0.9}
\caption{Lossless image compression performance of different lossless image codecs on DIV2K, CLIC.p, CLIC.m, Kodak, SCID, BRACS24 datasets in terms of bpsp $\downarrow$. For the bpsp, the lower ther better. The best performance in each dataset is bold.}
\label{tb:results_ll}
\centering
\small
\begin{tabular}{ccccccccc}
\toprule[1pt]
  Category & Codec     & Venue & DIV2K & CLIC.p & CLIC.m & Kodak & SCID & BRACS24 \tabularnewline
\midrule
    Classical & \begin{tabular}{@{}l@{}}PNG~\citep{boutell1997png} \\ JPEG-LS \cite{weinberger2000loco} \\ CALIC \cite{wu1997context} \\ JPEG2000 \cite{skodras2001jpeg} \\ WebP \cite{si2016research} \\ BPG \cite{yee2017medical} \\ FLIF \cite{sneyers2016flif} \\ JPEG-XL \cite{alakuijala2019jpeg} \end{tabular} 
    & \begin{tabular}{@{}c@{}}$-$ \\ TIP-2000 \\ TIP-1997 \\ $-$ \\ $-$ \\ $-$ \\ ICIP-2016 \\ $-$ \end{tabular} 
    & \begin{tabular}{@{}c@{}}4.23 \\ 2.99 \\ 3.07 \\ 3.12 \\ 3.11 \\ 3.28 \\ 2.91 \\ 2.88 \end{tabular} 
    & \begin{tabular}{@{}c@{}}3.93 \\ 2.82 \\ 2.87 \\ 2.93 \\ 2.90 \\ 3.08 \\ 2.72 \\ 2.63 \end{tabular} 
    & \begin{tabular}{@{}c@{}}3.93 \\ 2.53 \\ 2.59 \\ 2.71 \\ 2.73 \\ 2.84 \\ 2.48 \\ 2.36 \end{tabular} 
    & \begin{tabular}{@{}c@{}}4.35 \\ 3.16 \\ 3.18 \\ 3.19 \\ 3.18 \\ 3.38 \\ 2.90 \\ 2.87 \end{tabular} 
    & \begin{tabular}{@{}c@{}}1.79 \\ 2.11 \\ $-$ \\ 2.15 \\ 1.24 \\ 1.57 \\ $-$ \\ 1.26 \end{tabular} 
    & \begin{tabular}{@{}c@{}} 4.99 \\ 4.04 \\ $-$ \\ 3.83 \\ 3.94 \\ $-$  \\ $-$ \\ 3.67 \end{tabular}\tabularnewline
\midrule
    LIC & \begin{tabular}{@{}l@{}}L3C \cite{mentzer2019practical} \\ RC  \cite{mentzer2020learning} \\ iVPF \cite{zhang2021ivpf} \\ iFlow \cite{zhang2021iflow} \\ LLICTI \cite{kamisli2023learned} \\ ArIB-BPS \cite{zhang2024learned} \\ DLPR \cite{bai2024deep} \end{tabular} 
    & \begin{tabular}{@{}l@{}}CVPR-2019 \\ CVPR-2020 \\ CVPR-2021 \\ NeurIPS-2021 \\ TCSVT-2024 \\ CVPR-2024 \\ TPAMI-2024 \end{tabular} 
    & \begin{tabular}{@{}c@{}}3.09 \\ 3.08 \\ 2.68 \\ 2.57 \\ 2.77 \\ 2.55 \\ 2.55 \end{tabular} 
    & \begin{tabular}{@{}c@{}}2.94 \\ 2.93 \\ 2.54 \\ 2.44 \\ 2.79 \\ $-$ \\ 2.38 \end{tabular} 
    & \begin{tabular}{@{}c@{}}2.64 \\ 2.54 \\ 2.39 \\ 2.26 \\ $-$ \\ $-$ \\ 2.16 \end{tabular} 
    & \begin{tabular}{@{}c@{}}3.26 \\ $-$ \\ $-$ \\ $-$ \\ 2.99 \\ $-$ \\ 2.86 \end{tabular} 
    & \begin{tabular}{@{}c@{}}2.67 \\ $-$ \\ $-$ \\ $-$ \\ $-$ \\ $-$ \\ 1.58 \end{tabular}
    & \begin{tabular}{@{}c@{}}3.98 \\ $-$ \\ $-$ \\ $-$ \\ $-$ \\ $-$ \\ 3.61 \end{tabular} \tabularnewline
\midrule
    LLM & \begin{tabular}{@{}l@{}}\citet{deletang2024language} \\ $\mathbf{P}^{2}\mathbf{LLM}$ \end{tabular} 
    & \begin{tabular}{@{}l@{}}ICLR-2024 \\ This paper \end{tabular} 
    & \begin{tabular}{@{}c@{}} 4.17\\ \textbf{2.51} \end{tabular} 
    & \begin{tabular}{@{}c@{}} 3.89 \\ \textbf{2.35} \end{tabular} 
    & \begin{tabular}{@{}c@{}}3.76 \\ \textbf{2.08} \end{tabular} 
    & \begin{tabular}{@{}c@{}}3.96 \\ \textbf{2.83} \end{tabular} 
    & \begin{tabular}{@{}c@{}} 1.67 \\  \textbf{1.21} \end{tabular}
    & \begin{tabular}{@{}c@{}} 4.12 \\  \textbf{3.33} \end{tabular}\tabularnewline
\bottomrule[1pt]
\end{tabular}
\label{main results}
\end{table*}

Finally, the softmax function $\sigma(\cdot)$ is utilized to normalize these logits into a probability distribution, which equals a generalization of the logistic function to multiple dimensions:
\begin{equation}
    p(\mathbf{y}_{c}) = \sigma(\mathbf{y}_{c}) = \frac{e^{y_{c}^{z}}}{\sum_{z=0}^{255}e^{y_{c}^{z}}}.
\end{equation}
We can use encoded pixel values and corresponding probability density functions to conduct arithmetic coding. 

 \subsection{Fine-tuning and Implementation Details}\label{3.4}
  \textbf{Fine-tuning using Low-rank Adaptation.} To enhance the ability of next-pixel prediction of LLM and obtain optimum posterior distribution (as discussed in Cor. \ref{coro}), it is necessary to fine-tune the LLM using low-rank adaptation (LoRA)~\citep{hu2021lora}. LoRA enables the LLM to adapt to a customized task in a computationally efficient manner~\citep{li2023loftq}. By following \citet{deletang2024language}, we mainly explore the effectiveness of language models for lossless image compression. To this end, the Llama 3 series~\citep{dubey2024llama}, open-source LLMs released by Meta, are used. We utilize the pre-trained Llama 3 series 8B base model (The effect of other model sizes is presented in \textcolor{red}{Appendix}) provided by Huggingface\footnote{https://huggingface.co/meta-llama}. \\
 \textbf{Patch Processing and Parallel Accelerating.} By following \citet{deletang2024language}, we split the overall image into sequential non-overlapped patches for compression.  Meanwhile, different patches can be independently processed in a batch manner, which enables parallel accelerating. 
\section{Experiments}

\textbf{Datasets.} By following previous works~\citep{bai2024deep}, seven different datasets are imposed to evaluate the lossless compression performance of different approaches. We follow \citet{bai2024deep} to use the DIV2K high-resolution training dataset~\citep{agustsson2017ntire} for fine-tuning the LLM, where each image is cropped into non-overlapped patches. The details of adopted datasets can be found in Table \ref{dataset} and \textcolor{red}{Appendix}. 

\textbf{Training Details.} As the rationale of LoRA is to approximate a large matrix by two low-rank decomposed matrixes, the rank and corresponding alpha coefficient in LoRA would significantly affect the performance. We ablate the rank in some predefined values, and the alpha coefficient is twice as much as the rank for a defaulted setting. After ablation analysis (in \textcolor{red}{Appendix}), the rank and alpha coefficient are set to 64, and 128, respectively. The target modules of LoRA include query, key, value, and output projections. The patch size determines the length of context information for LLM. We ablate different patch sizes in predefined values and choose the size of $16\times16$. The task prompt of LLM is used as described in sec. \ref{3.1} (The effect of other task prompts is presented in \textcolor{red}{Appendix}). The initial rate of the cosine decay learning scheduler is set to $1\times10^{-4}$ with a warming-up of 1000 steps. We use 4 NVIDIA A800 GPUs for fine-tuning with a batch size of 8 per GPU.

\textbf{Baselines.} To evaluate the effectiveness of our proposed method, various baseline codes are introduced as follows:
1) \textbf{Classical Codes.} Classical codes usually compress the image using handcrafted priors and elaborated framework designs. Here, we use some widely-adopted classical codes, including PNG~\citep{boutell1997png}, JPEG-LS~\citep{weinberger2000loco}, CALIC~\citep{wu1997context}, JPEG2000~\citep{skodras2001jpeg}, WebP~\citep{si2016research}, BPG~\citep{yee2017medical}, FLIF~\citep{sneyers2016flif}, and JPEG-XL~\citep{alakuijala2019jpeg}.
2) \textbf{Learned Image Compression (LIC).} LIC models usually directly minimize the rate cost by deep neural networks. In this branch, residual coding-based pipelines have achieved SOTA compression performance, where the residual information of lossy compression is compressed by arithmetic coding. We utilize some SOTA LIC models for comparison, including L3C~\citep{mentzer2019practical}, RC~\citep{mentzer2020learning}, iVPF~\citep{zhang2021ivpf}, iFlow~\citep{zhang2021iflow}, LLICTI~\citep{kamisli2023learned}, ArIB-BPS~\citep{zhang2024learned}, and DLPR~\citep{bai2024deep}. 
3) \textbf{LLM-based Compressor.} We mainly reproduce \citet{deletang2024language}' method to compress the RGB images by maintaining their key components, including using a pre-trained LLM, proxy tokens, and a channel-independent manner. Practically, we discard the proxy tokens to use the same tokenization manner adopted by ours, as it may be more sophisticated to ensure a lossless compression due to the appended bit for lossy tokenization.
\begin{table}[!t]
\renewcommand\arraystretch{1}
\centering
\caption{Ablation Study. Channel-Indep. means RGB images are compressed in a channel-independent manner for LLM. Channel-Corre. means RGB images are compressed by next-pixel prediction, as proposed in sec. \ref{3.1} (see Figure \ref{fig:flatten}). FT denotes the fine-tuning. w/: With and w/o: Without. Task prompt for channel-independent setting is \textit{``R/G/B channel of a flattened RGB image. Predict the next sub-pixel based on previous sub-pixels.''}  }
\begin{adjustbox}{max width=0.5\textwidth}
\begin{tabular}{c|cc|ccc}
\hline
                                     & \multicolumn{4}{c|}{Pixel Prediction Chat Template}                                                                                              & \multirow{3}{*}{bpsp$\downarrow$} \\ \cline{1-5}
                                     & \multicolumn{2}{c|}{Pixel-level Prior}                       & \multicolumn{2}{c|}{In-context Learning}                                         &                       \\ \cline{2-5}
                                     & \multicolumn{1}{l|}{Channel-Indep.} & \multicolumn{1}{l|}{Channel-Corre.} & \multicolumn{1}{l|}{w/o Task Prompt} & \multicolumn{1}{l|}{w/ Task Prompt} &                       \\ \hline
\multirow{4}{*}{w/o FT} & \cellcolor{gray!20}{\checkmark}                 &                 \(\times\)                        &  \cellcolor{gray!20}{\checkmark}                                   &                   \(\times\)                       &       4.55                \\ \cline{2-6}
                                     &  \cellcolor{gray!20}{\checkmark}                 &             \(\times\)                            &                  \(\times\)                     &  \cellcolor{gray!20}{\checkmark}                                      &            4.46           \\ \cline{2-6} \cline{2-6}
                                    &           \(\times\)           &  \cellcolor{gray!20}{\checkmark}                                     &  \cellcolor{gray!20}{\checkmark}                                   &                      \(\times\)                    &      3.96                 \\ \cline{2-6} \cline{2-6}
                                     &           \(\times\)          &  \cellcolor{gray!20}{\checkmark}                                     &             \(\times\)                          &  \cellcolor{gray!20}{\checkmark}                                      &    3.83                   \\ \hline \hline
\multirow{4}{*}{w/ FT} & \cellcolor{gray!20}{\checkmark}                 &                 \(\times\)                        &  \cellcolor{gray!20}{\checkmark}                                   &                   \(\times\)                       &       3.95                \\ \cline{2-6}
                                     &  \cellcolor{gray!20}{\checkmark}                 &             \(\times\)                            &                  \(\times\)                     &  \cellcolor{gray!20}{\checkmark}                                      &            3.76           \\ \cline{2-6} \cline{2-6}
                                    &           \(\times\)           &  \cellcolor{gray!20}{\checkmark}                                     &  \cellcolor{gray!20}{\checkmark}                                   &                      \(\times\)                    &      2.99                 \\ \cline{2-6} \cline{2-6}
                                     &           \(\times\)          &  \cellcolor{gray!20}{\checkmark}                                     &             \(\times\)                          &  \cellcolor{gray!20}{\checkmark}                                      &    \textbf{2.83}                   \\ \hline
\end{tabular}
\end{adjustbox}
\label{ablation 1}
\vspace{-0.3cm}
\end{table}
\subsection{Main Results}
We evaluate the lossless compression performance of different codes using bit-per-subpixel (bpsp). For the bpsp, the lower the better. As illustrated in Table \ref{main results}, our proposed P$^{2}$-LLM achieves the best performance in all datasets, compared all classical and LIC baselines with an obvious margin. For example, P$^{2}$-LLM achieves 2.08 and 2.83 bpsp, suppressing the best LIC approach (DLPR) with 2.16 and 2.86 bpsp. Meanwhile,  P$^{2}$-LLM  beats the best classical compressor, JPEG-XL. Note that \citet{deletang2024language} approach only outperforms the PNG, which is reasonable as they cannot generalize to widely-used images (\textit{e.g.,} RGB images) due to the lack of effective pixel-level semantic context and the fine-tuning. Instead, our proposed P$^{2}$-LLM awakes LLM's compression ability by various designs. 
\begin{figure}[!t]
    \centering
    \begin{subfigure}[b]{0.235\textwidth}
        \centering
        \includegraphics[width=\textwidth]{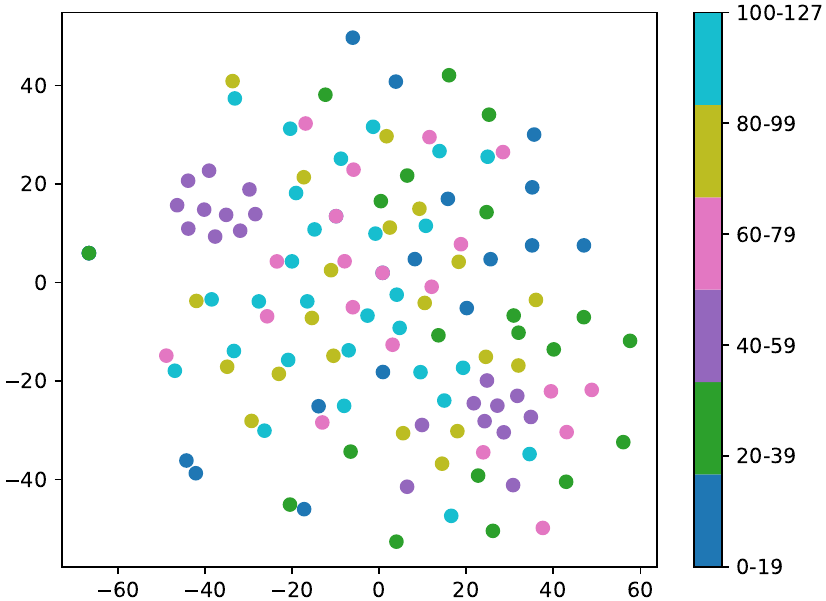}
        \caption{ASCII Tokens}
        \label{fig:sub1}
    \end{subfigure}
    \hfill
    \begin{subfigure}[b]{0.235\textwidth}
        \centering
        \includegraphics[width=\textwidth]{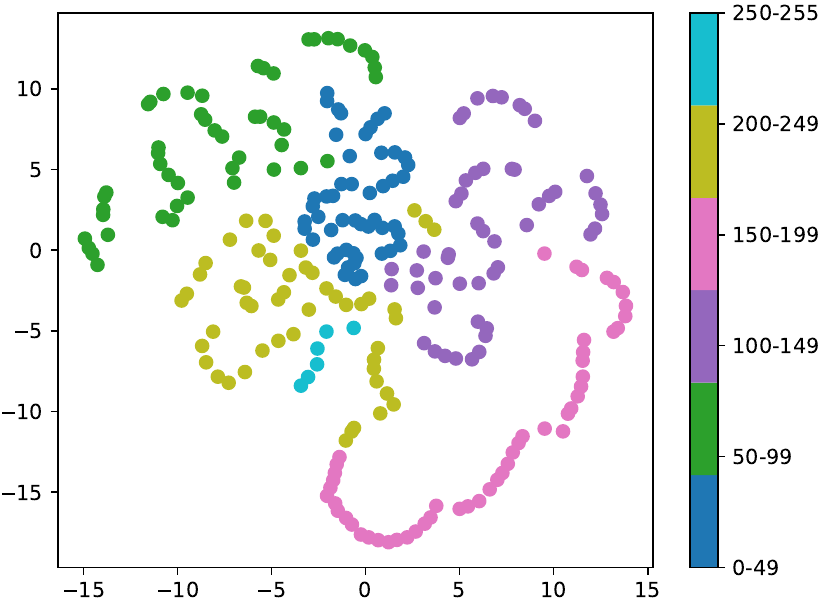}
        \caption{Numerical Tokens}
        \label{fig:sub2}
    \end{subfigure}
    \caption{Token embedding visualization of pixel values using t-SNE (Note that the dimension of each token embedding is $4096$ in Llama 3 series). (a) Pixel values are tokenized by proxy tokens, as in \citet{deletang2024language} (adopt a data remapping from [0,255] to [0,127]). (b) Using digital token dictionary as in sec. \ref{3.2}.  }
    \label{fig:t-sne}
\end{figure}
\begin{table}[!t]
\centering
\caption{Comparison of different lengths of the context for LLM. The token number of the task prompt is 19 for all settings.}
\renewcommand\arraystretch{0.85}
\begin{adjustbox}{max width=0.43\textwidth}
\begin{tabular}{ccl}
\hline
Patch Size  & \#Total Tokens of the Context & bpsp \\ \hline
8$\times$8$\times$3                                               & 211                                    &       3.10        \\
\rowcolor{gray!20}12$\times$12$\times$3                                               & 451                                    &       2.95        \\
16$\times$16$\times$3                                            & 787                                    & \textbf{2.83}         \\
\rowcolor{gray!20} 24$\times$24$\times$3                                             & 1747                                    &     2.99          \\
32$\times$32$\times$3                                           & 3091                                   &   3.49            \\ \hline
\end{tabular}
\end{adjustbox}
\vspace{-0.2cm}
\label{different length}
\end{table}
\begin{table}[!t]
\renewcommand\arraystretch{0.85}
\centering
\caption{Comparison of runtime (second/image) on Kodak dataset.}
\begin{adjustbox}{max width=0.43\textwidth}
\begin{tabular}{ccc}
\hline
Codec & Encoding Time & Decoding Time \\ \hline
JPEG-XL        & 0.73                   & 0.08                   \\
BPG            & 2.38                   & 0.13                   \\
L3C            & 8.17                   & 7.89                   \\
DLPR           & 1.26                   & 1.80                   \\ \hline
\rowcolor{gray!15} \citet{deletang2024language}           & 15.13                   & 272.05               \\
\rowcolor{gray!15} P$^{2}$-LLM           &     14.89               &   273.11             \\\hline
\end{tabular}
\end{adjustbox}
\label{runtime}
\begin{minipage}{0.43\textwidth}
\vspace{1mm}
\footnotesize{Note: 8 NVIDIA A800 GPUs (\#batch size: 16, \#subprogress: 2 per GPU) for parallel processing. P$^{2}$-LLM and \citet{deletang2024language}' model adopt the same context for fairness, leading to a similar runtime.}
\end{minipage}
\vspace{-0.5cm}
\end{table}
\subsection{Detailed Analysis of Each Key Component}
We carefully analyze the effectiveness of each key component using the ablation study on the Kodak testing dataset. Some interesting conclusions can be presented as follows. \\
\textbf{-- Fine-tuning (as discussed in sec. \ref{3.4}) \textcolor{red}{{\textit{cannot}}} fully awake LLM's compression ability.} As illustrated in Table \ref{ablation 1}, it can be observed that the fine-tuning can significantly improve the compression performance under the same setting, which is reasonable as the LLM's understanding ability increases a lot for pixel sequences, resulting in more accurate next-pixel predictions. However, it should be noted that simply fine-tuning LLM cannot result in SOTA compression performance. For example, a 3.76 bpsp score (last-third row) is achieved with channel-independent and task prompt settings. Such performance still has a significant downside compared SOTA models as in Table \ref{main results}.\\
\textbf{-- Pixel-level priors and In-context learning (as discussed in sec. \ref{3.1}) are important catalysts, especially the former.} As illustrated in Table \ref{ablation 1}, pixel-level priors and in-context learning can improve the compression performance, regardless of with or without fine-tuning settings. Especially, without fine-tuning LLM, we can observe that simple usage of channel correlations can intensively increase the performance, $\textit{i.e.,}$ from \textbf{4.55} (first row) to \textbf{3.96} (third row) bpsp. This is reasonable as the LLM can leverage the intra-pixel inter-channel correlations for more accurate next-pixel reasoning. Meanwhile, it seems that the task prompt can moderately enhance the understanding of LLM for pixel sequence with better compression performance using the context, \textit{e.g.,} from 3.96 (third row) to 3.83 (fourth row) bpsp.  \\\textbf{-- Two-step lossless pixel tokenization (as discussed in sec. \ref{3.2}) can maintain pixel-level semantic context with more compact representations.} As shown in Figure \ref{fig:t-sne}, we visualize the token embeddings of pixel values from 0 to 255. These embeddings are queried from the \texttt{embed\_tokens} layer of LLM. We can observe that the pixels with closer values are roughly closer in feature space when our numerical tokens are adopted, thus pixel-level semantic context can be preserved in language space. This aids LLM in understanding the relationship between pixels better, compared with ASCII-based tokens. \\
\textbf{-- Predictive distribution sampling (as discussed in sec. \ref{3.3}) results in accurate and compact probability representation.} As shown in Figure 
\ref{fig:sampling},  predictive logtis between 0 and 255 are presented in the middle column. For each prediction, the peak of those logits is roughly around the encoded pixel value, which showcases that LLMs can effectively predict the next pixel by understanding the relationship between pixel values. After the softmax function, the probability representation is extremely compact, which can benefit the AC with better compression performance.\\
\textbf{-- Moderate patch size is the best.} As illustrated in Table \ref{different length}, the length of context information can affect the compression performance, where a moderate patch size is the best. This may be reasonable as modern LLMs usually struggle with long-context scenarios due to the degradation of position embeddings~\citep{li2024long}.

\section{Limitation and Conclusion}
\textbf{Limitation.} Although we have observed that the LLM-based compressor can beat classical and LIC-based codes using its unprecedented intelligence, its decoding time as illustrated in Table \ref{runtime} is much slower than other baselines due to the inherent downside of autoregressive models~\citep{kizhakkumkara2023autoregressive}. More investigations about balancing effectiveness and efficiency will be explored in the future.

In this paper, to fully utilize LLMs' intelligence for lossless image compression, we have introduced P$^{2}$-LLM, a next-pixel prediction-based LLM that leverages pixel-level priors, in-context abilities, pixel-level semantic preservation, and fine-tuning strategy to improve lossless image compression performance in language space. This mitigates the gap between theoretical and practical compression performance for LLM. Extensive experiments show that P$^{2}$-LLM can beat SOTA classical and learned lossless compressors with obvious gains.

\newpage
\newpage
\appendix

\end{document}